\documentclass[authorversion]{acmart13}
\usepackage{booktabs} 
\usepackage{multirow}

\usepackage[ruled]{algorithm2e} 

\setcopyright{rightsretained}

\acmDOI{0000001.0000001}

\begin{document}
\title{Stance Detection on Tweets: An SVM-based Approach}
\author{Dilek K\"u\c{c}\"uk}
\orcid{0000-0003-2656-1300}
\affiliation{
  \institution{T\"UB\.ITAK Energy Institute}
  \department{Electrical Power Technologies Department}
}
\email{dilek.kucuk@tubitak.gov.tr}

\author{Fazli Can}
\email{canf@cs.bilkent.edu.tr }
\affiliation{%
  \institution{Bilkent University}
  \department{Bilkent Information Retrieval Group, Computer Engineering Department}
}

\begin{abstract}
Stance detection is a subproblem of sentiment analysis where the stance of the author of a piece of natural language text for a particular target (either explicitly stated in the text or not) is explored. The stance output is usually given as \emph{Favor}, \emph{Against}, or \emph{Neither}. In this paper, we target at stance detection on sports-related tweets and present the performance results of our SVM-based stance classifiers on such tweets. First, we describe three versions of our proprietary tweet data set annotated with stance information, all of which are made publicly available for research purposes. Next, we evaluate SVM classifiers using different feature sets for stance detection on this data set. The employed features are based on unigrams, bigrams, hashtags, external links, emoticons, and lastly, named entities. The results indicate that joint use of the features based on unigrams, hashtags, and named entities by SVM classifiers is a plausible approach for stance detection problem on sports-related tweets.
\end{abstract}

\ccsdesc[500]{Information systems~Sentiment analysis}
\ccsdesc[500]{Information systems~Web and social media search}
\ccsdesc[500]{Computing methodologies~Language resources}

\keywords{Stance detection, Twitter, SVM, social media analysis, unigrams, named entities}

\thanks{Authors' addresses: D. K\"u\c{c}\"uk, Electrical Power Technologies Department, T\"UB\.ITAK Energy Institute, Ankara, Turkey; F. Can, Bilkent Information Retrieval Group, Computer Engineering Department, Bilkent University, Ankara, Turkey}

\maketitle

\renewcommand{\shortauthors}{D. K\"u\c{c}\"uk and F. Can}

\section{Introduction}\label{intro}
Stance detection is one of the subproblems in sentiment analysis (opinion mining) \cite{pang2008opinion} and is a considerably recent research area. In stance detection, the position of an author of a piece of text is explored for a particular target (an entity, event, idea etc.) which may be explicitly stated in the text or not. The stance output is usually expected from the set: \{\emph{Favor}, \emph{Against}, \emph{Neither}\} \cite{mohammad2016semeval} while in the sentiment analysis task, the output is usually one of \emph{Positive}, \emph{Negative}, or \emph{Neutral} and usually no sentiment target is considered.

In this paper, we present our SVM-based stance detection approaches on sports-related tweets\footnote{A preliminary version of this paper has been presented in \cite{kucuk2017stance}. The current study also includes our previous experiments published as a preprint in \cite{Kucuk2017joint}.}. The main contributions of this study are presented below:
\begin{itemize}
\item   Three different versions of a tweet data set in Turkish annotated with stance information are presented. The corresponding annotations are made publicly available for research purposes. The second and third versions of the data set are described in the current paper for the first time. They are the extended versions of a previously proposed data set \cite{kucuk2017stance} and the extended versions have been annotated by two annotators as opposed to the initial version which was annotated by a single annotator. Additionally, the number of tweets in the final version of the data set is more than 1.5 times the number of tweets in the initial version. The tweets in the data sets are about two popular football clubs which are, hence, the stance targets. To the best of our knowledge, the data sets are the first stance-annotated resources for Turkish, two of which are annotated by two native speakers, and they constitute the first sport-related stance-annotated resources in general.
\item   We provide the results of our experiments based on SVM classifiers (one for each target) on the data set versions using features based on unigrams, bigrams, hashtags, external links, and emoticons. Named entities in tweets are also considered as features for stance classification and the results of the related experiments are provided as well. These experiments form plausible baselines with which future work on stance detection can be compared.
\end{itemize}

The rest of the paper is organized as follows: In Section \ref{related}, a review of the literature on stance detection is provided. In Section \ref{dataset}, we describe three different versions of our tweet data set annotated with the stance target and stance information. Section \ref{svm} includes the details of our SVM-based stance classifiers and their evaluation results with discussions. Future research topics based on the current study are provided in Section \ref{future}. Finally, Section \ref{conc} concludes the paper with a summary of main points.

\section{Related Work}\label{related}

Literature on stance detection includes \cite{somasundaran2010recognizing} where a stance detection approach was presented, based on sentiment and arguing features, along with an arguing lexicon automatically compiled. This approach was reported to perform better than baseline systems which were distribution-based and unigram-based systems \cite{somasundaran2010recognizing}. In studies such as \cite{walker2012stance,walker2012your}, it was concluded that considering the dialog structure of online debate posts improved stance classification performance on these posts.

In \cite{hasan2013stance}, stance detection experiments were performed using machine learning algorithms, training data sets, features, and inter-post constraints in on-line debates, and useful conclusions based on these experiments were obtained. It was found out that for stance detection, sequence models like HMMs performed better than non-sequence models like Naive Bayes (NB) \cite{hasan2013stance}. In another study \cite{misra2013topic}, it was pointed out that topic-independent features could be used for disagreement detection in on-line dialogues. The features utilized include agreement, cue words, denial, hedges, duration, polarity, and punctuation \cite{misra2013topic}. In \cite{faulkner2014automated}, stance detection on student essays was considered. After using linguistically-motivated feature sets with multivalued NB and SVM as the learning models, the authors reported that they outperformed their baseline approaches \cite{faulkner2014automated}. In another related work \cite{dori2015controversy}, it was claimed that Wikipedia could be used to determine stances on controversial topics. A unified framework for stance classification using probabilistic modeling of online debate forums was proposed in \cite{sridhar2015joint}.

Considering more recent related work, in \cite{augenstein2016stance}, bidirectional conditional encoding was employed for stance detection for unseen targets. It is reported that this approach achieved state-of-the art performance rates \cite{augenstein2016stance} on \emph{SemEval 2016 Twitter Stance Detection Corpus} \cite{mohammad2016semeval}. A stance-community detection approach, called SCIFNET, was presented in \cite{chen2016scifnet}. The approach created networks of people who were stance targets, automatically from document collections using stance expansion and refinement techniques \cite{chen2016scifnet}. A tweet data set annotated with stance information regarding six predefined targets was described in \cite{mohammad2016dataset} where crowdsourcing was employed to annotate the data set. This set was also annotated with sentiments, so it could help reveal relationships between stance and sentiment information \cite{mohammad2016dataset}. An extension to this study was presented in \cite{mohammad2017stance} where further stance and sentiment classification experiments were conducted on the data set annotated with both information.

In the previously mentioned study \cite{mohammad2016semeval}, SemEval 2016's shared task on \emph{Twitter Stance Detection} was described in details. The study also presented the evaluation results of 19 systems participating in two subtasks (one subtask with a training data set provided and another subtask without an annotated data set) of the shared task \cite{mohammad2016semeval}. The system performing the best in the first task was presented in \cite{zarrella2016mitre} which employed recurrent neural networks. It achieved an average F-Measure of 67.8\% for this task. Another system that also performed well for this task of the competition was presented in \cite{tutek2016takelab}. An ensemble of genetic algorithms was used for the stance detection task \cite{tutek2016takelab}. A system that performed well for both tasks of the competition was presented in \cite{wei2016pkudblab} which was based on a convolutional neural network. Interested readers are referred to \cite{mohammad2016semeval} for references to the systems participating in this stance annotation competition.

In \cite{ebrahimijoint2016}, a log-linear model for stance classification in tweets was proposed where the interactions between the stance target, stance, and sentiment were modeled. In \cite{gadek2017}, the authors showed that extracting and using contextonyms ("contextually related words") helped improve stance detection in tweets. In \cite{sobhani2017dataset}, a data set for multi-target stance detection was presented together with experiments on this data set.

Another recent topic closely related to stance detection is argumentation (or, argument) mining. The aim of the argumentation mining task is to identify the particular arguments, related components, and relations in natural language texts \cite{nguyen2015extracting}. These texts are usually in the form of on-line debates, legal documents, and student essays \cite{nguyen2015extracting}. There are also studies that performed joint argument mining and stance detection \cite{sobhani2015argumentation}.

\section{Turkish Tweet Data Sets Annotated with Stance Information}\label{dataset}
In the preliminary version of our current study \cite{kucuk2017stance}, tweets about popular sports clubs were considered for stance detection where the targets were determined to be \emph{Galatasaray} (namely, \emph{Target-1}) and \emph{Fenerbah\c{c}e} (namely, \emph{Target-2}) which are two popular football clubs in Turkey. The tweets in the stance data set was obtained from the randomly compiled one million tweets published between August 18 and September 6, 2015 within the course of a public health surveillance study \cite{Kucuk2017}. During annotation, only the stance classes of \emph{Favor} and \emph{Against} were considered, therefore, there was no tweet instance labelled with the \emph{Neither} class \cite{kucuk2017stance}. In most of the tweets, the target is exactly the sports club in its entirety, while in some other, either the management of the club or some footballers are criticized or praised. In all of these cases, the club is determined to be the stance target. The corresponding stance-annotated data set contains 700 tweets, where 175 tweets are in favor of and 175 tweets are against \emph{Target-1}, and similarly 175 tweets are in favor of and 175 are against \emph{Target-2}. The stance annotations for this initial version of the data set were publicly shared at \texttt{http://ceng.metu.edu.tr/}$\sim$\texttt{e120329/} \texttt{Turkish\_Stance\_Detection\_Tweet\_Dataset.csv}. The stance annotations on this data set were carried out by a single annotator who is a native speaker of Turkish \cite{kucuk2017stance}.

Within the course of the current study, another annotator, who is again a native speaker of Turkish, is provided with the initial data set of 700 tweets and a small set of guidelines, and these tweets have been annotated with stance information by this second annotator, independently. The percentage of matching annotations for the tweets about \emph{Target-1} is found to be 98.86\% and that for the tweets on \emph{Target-2} is found as 97.14\%. These numbers are calculated by summing the number of tweets annotated with the \emph{Favor} class by both annotators and the number of tweets annotated as \emph{Against} by both annotators, and then diving the summation to total number of tweets. Hence, the percentage of matching annotations for the overall data set is calculated as 98\%, i.e., the two annotators have indicated the same stance for 686 of the 700 tweets in the data set. For these two annotations, Cohen's kappa for inter-annotator agreement is found to be 96\% which is considered as "a very good agreement"\footnote{The formula for calculating Cohen's kappa is $\frac{P_o - P_e}{1 - P_e}$ \cite{cohen1960coefficient}. In this formula, $P_o$ is the relative observed agreement among annotators (0.98 in our case) and $P_e$ is the hypothetical probability of chance agreement (0.5 in our case).}. This high level of inter-annotator agreement can be attributed to the particular domain of the tweets considered, i.e., stance annotations for sports clubs may have led higher levels of inter-annotator agreement compared to the corresponding stance annotations for other text domains. Hence, we have created a second version of our stance-annotated data set in which only 686 tweets agreed by both annotators are included. Of these 686 tweets, 173 are in favor and 173 are against \emph{Target-1} while 173 are in favor and 167 are against \emph{Target-2}. Similar to its initial version, this second version of the stance-annotated tweet data set is publicly shared at \texttt{http://ceng.metu.edu.tr/}$\sim$\texttt{e120329/} \texttt{Turkish\_Stance\_Detection\_Tweet\_Dataset\_V2.csv}.

In order to increase the size of this data set, 400 new tweets have been extracted from the initial randomly-compiled tweet set of one million tweets. The first annotator has marked 100 of them as in favor of and 100 of them against \emph{Target-1}, and similarly 100 of them as in favor of and 100 of them against \emph{Target-2}. Following the same annotation scheme, the second annotator has also been asked to annotate these 400 tweets with stance information. It is found that 379 tweets are marked with the same stance class by both annotators. This tweet set of agreed 379 tweets has been added to the second version of the aforementioned tweet set and hence we have arrived at the third version of the stance-annotated tweet set of 1,065 tweets. The percentage of matching annotations performed by both annotators is 96.81\% and Cohen's kappa for the inter-annotator agreement on this final data set is found to be 93.6\% which is again considered as "a very good agreement". Of all tweets in this final data set, 269 of them are in favor of and 268 of them are against \emph{Target-1} while 269 of them are in favor of and 259 of them are against \emph{Target-2}. The annotations for this final and extended data set is made publicly-available at \texttt{http://ceng.metu.edu.tr/}$\sim$\texttt{e120329/} \texttt{Turkish\_Stance\_Detection\_Tweet\_Dataset\_V3.csv}.

Summary information regarding all three versions of the stance-annotated tweet data set is provided in Table \ref{tab:datasetsummary}.

\begin{table}[h!]
\centering
\caption{A Summary of the Three Versions of the Stance-Annotated Tweet Data Set}
\label{tab:datasetsummary}
\begin{tabular}{|c|c|c|c|c|c|c|}
\hline
&&\multicolumn{5}{|c|}{\# of Tweets Annotated} \\
\cline{3-7}
&&\multicolumn{2}{|c|}{For Target-1} & \multicolumn{2}{|c|}{For Target-2} & \multirow{2}{*} {\textbf{TOTAL}} \\
\cline{3-6}
Stance Data Set & \# of Annotators & Favor & Against & Favor & Against & \\
\hline
Version-1& 1 & 175 & 175 & 175 & 175 & \textbf{700} \\
\hline
Version-2& 2 & 173 & 173 & 173 & 167 & \textbf{686} \\
\hline
Version-3& 2 & 269 & 268 & 269 & 259 & \textbf{1,065} \\
\hline
\end{tabular}
\end{table}

Below provided are four sample tweets from the final version of the data set, together with their translations in English, and the corresponding stance-annotations, for illustrative purposes:\\\\
Original Tweet: \hspace{3px} \texttt{ve biz iyi ki Galatasarayl{\i}y{\i}z}\\
Translation: \hspace{15px} \texttt{and fortunately we are supporters of Galatasaray}\\
Target: \hspace{33px} \texttt{Galatasaray}\\
Stance: \hspace{33px} \texttt{Favor}\\\\
Original Tweet: \hspace{3px} \texttt{Bu grup ha\c{s}lar Galatasaray{\i} :D}\\
Translation: \hspace{15px} \texttt{This group will boil Galatasaray :D}\\
Target: \hspace{33px} \texttt{Galatasaray}\\
Stance: \hspace{33px} \texttt{Against}\\\\
Original Tweet: \hspace{3px} \texttt{Fenerbah\c{c}eli olmak ayr{\i}cal{\i}kt{\i}r...}\\
Translation: \hspace{15px} \texttt{It is a privilege to be the supporter of Fenerbah\c{c}e...}\\
Target: \hspace{33px} \texttt{Fenerbah\c{c}e}\\
Stance: \hspace{33px} \texttt{Favor}\\\\
Original Tweet: \hspace{3px} \texttt{Kanser olmaya haz{\i}r m{\i}y{\i}z ? \#Fenerinma\c{c}{\i}var}\\
Translation: \hspace{15px} \texttt{Ready to get cancer ? \#Fenerhasamatch}\\
Target: \hspace{33px} \texttt{Fenerbah\c{c}e}\\
Stance: \hspace{33px} \texttt{Against}\\

These three versions of the data set constitute the first publicly available stance-annotated data sets for Turkish where the second and the third ones have been created by two annotators. Hence, they are significant resources due to the scarcity of annotated data sets, linguistic resources, and NLP tools available for Turkish. Additionally, to the best of our knowledge, they are also significant for being the initial stance-annotated data sets including sports-related tweets, as previous stance detection data sets mostly include on-line debates on political\slash ethical issues.

\section{Stance Detection Experiments Using SVM Classifiers}\label{svm}

\begin{figure}
\center \scalebox{0.75}
  {\includegraphics{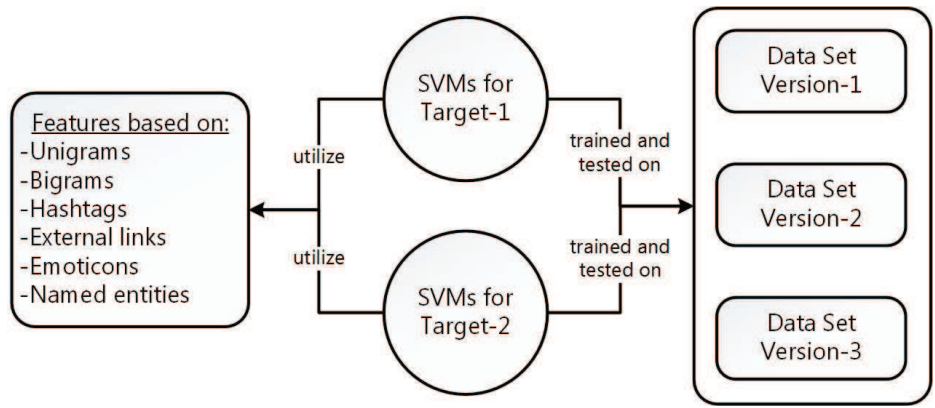}}
  \caption{An Overview of the Stance Detection Experiments Conducted.}
  \label{fig:overview}
\end{figure}

In this section, we present the details of the stance detection experiments using SVM classifiers on the tweet data set versions described in the previous section. A schematic representation of these experiments is presented in Figure \ref{fig:overview} and the details of them are provided in the following subsections.

\subsection{Experiments Using Ngrams and Hashtags as Features}\label{experimentsTr1}

It is emphasized in the related literature \cite{somasundaran2010recognizing} that unigram-based methods are reliable for the stance detection task and similarly unigram-based models have been used as baseline models in studies such as \cite{mohammad2016semeval}.

In order to be used as a baseline and reference system for further studies on stance detection in Turkish tweets, we have trained two SVM classifiers (one for each target) using unigrams as features. Before the extraction of unigrams, we have employed automated preprocessing to filter out the stopwords in tweets. The stopword list used is the list presented in \cite{kucuk2011exploiting} which, in turn, is the slightly extended version of the stopword list provided in \cite{can2008information}.

We have used the SVM implementation available in the Weka data mining application \cite{weka2009} where this particular implementation employs the SMO algorithm \cite{platt1999} to train a classifier with a linear kernel. The 10-fold cross-validation results of the two classifiers on the three versions of the data set are provided in Table \ref{tab:result1} using the metrics of precision (P), recall (R), and F-Measure (F).

\begin{table}[h!]
\centering
\caption{Evaluation Results of the Unigram-based SVM Classifiers}
\label{tab:result1}
\begin{tabular}{|l|l|c|c|c|c|c|c|c|c|c|}
\hline
&&\multicolumn{9}{|c|}{Stance Data Set} \\
\cline{3-11}
\multirow{3}{*}{Target}&\multirow{3}{*}{Class}&\multicolumn{3}{|c|}{Version-1}&\multicolumn{3}{|c|}{Version-2}&\multicolumn{3}{|c|}{Version-3} \\
\cline{3-11}
&& P (\%) & R (\%) & F (\%) & P (\%) & R (\%) & F (\%) & P (\%) & R (\%) & F (\%) \\
\hline
\multirow{3}{*}{Target-1} & Favor & 75.2 & 92.0 & 82.8 & 73.0 & 89.0 & 80.2 & 73.3 & 94.1 & 82.4 \\
& Against & 89.7 & 69.7 & 78.5 & 85.9 & 67.1 & 75.3 & 91.7 & 65.7 & 76.5 \\
& \textbf{Average} & \textbf{82.5} & \textbf{80.9} & \textbf{80.6} & \textbf{79.5} & \textbf{78.0} & \textbf{77.8} & \textbf{82.5} & \textbf{79.9} & \textbf{79.5} \\
\hline
\multirow{3}{*}{Target-2} & Favor & 68.5 & 83.4 & 75.3 & 69.7 & 89.0 & 78.2 & 75.2 & 90.0 & 81.9 \\
& Against & 78.8 & 61.7 & 69.2 & 84.0 & 59.9 & 69.9 & 86.9 & 69.1 & 77.0 \\
& \textbf{Average} & \textbf{73.7} & \textbf{72.6} & \textbf{72.2} & \textbf{76.7} & \textbf{74.7} & \textbf{74.1} & \textbf{80.9} & \textbf{79.7} & \textbf{79.5} \\
\hline
\end{tabular}
\end{table}

The performance evaluation results in Table \ref{tab:result1} are very promising for both targets, considering the fact that they are part of the initial experiments on the data sets. Although the results on the first version of the data set are particularly higher for \emph{Target-1}, the gap between the F-Measure rates of the targets disappears on the third version. That is, the performance rates of the unigram-based classifier for \emph{Target-2} increases considerably during the experiments on the second and third versions of the data set. We observe some slight performance decreases of the classifier for \emph{Target-1} when tested on the second and third versions of the data set. The decrease is even lower on the third version of the data set. The performance decrease for \emph{Target-1} on the second data set may be due to the peculiarities of the unigrams in the excluded tweets from the first version of the data set to arrive at the second one. On the third data set, the performance rates of the classifiers for both targets converge to 79.5\% in terms of F-Measure and are quite favorable.

It is observed that the performance rates of both classifiers (in terms of F-Measure) are better for the \emph{Favor} class for both targets when compared with the corresponding results for the \emph{Against} class. This outcome may be due to the common use of some terms when expressing positive stance towards sports clubs in Turkish tweets. The same percentage of common terms may not have been observed in tweets during the expression of negative stances towards the targets. Yet, completely the opposite pattern is observed in stance detection results of baseline systems given in \cite{mohammad2016semeval}, i.e., better F-Measure rates have been obtained for the \emph{Against} class when compared with the \emph{Favor} class \cite{mohammad2016semeval}. Some of the baseline classifiers in \cite{mohammad2016semeval} are SVM-based systems using unigrams and ngrams as features similar to our study, but their data sets include all three stance classes of \emph{Favor}, \emph{Against}, and \emph{Neither}, while our data set comprises tweets classified only as \emph{Favor} or \emph{Against}. Another difference is that the data sets in \cite{mohammad2016semeval} have been divided into training and test sets, while in our study we provide 10-fold cross-validation results on the whole data set. Additionally, the domains of the data sets used in \cite{mohammad2016semeval} are on political\slash social issues while our data sets contain tweets about sports clubs.

On the other hand, we should note that SVM-based sentiment analysis approaches (such as \cite{poursepanj2013uottawa}) have been reported to achieve better F-Measure rates for the \emph{Positive} sentiment class when compared with the results for the \emph{Negative} class. Hence, our performance evaluation results for each stance class seem to be in line with those of such sentiment analysis systems. Yet, further experiments on larger and diverse data sets can be conducted and the results achieved can again be compared with the stance detection results given in the literature.

We have also evaluated the performance of SVM classifier versions which use only bigrams as features, since ngram-based classifiers have been reported to perform favorably for the stance detection problem \cite{mohammad2016semeval}. However, we have observed that using bigrams as the sole features of the SVM classifiers leads to quite poor results. Our experiments indicate that unigram-based features lead to superior results compared to the results obtained using bigrams as features. Yet, ngram-based features may be employed on the prospective stance data sets from other domains to verify this conclusion within the course of future work.

In order to observe the contribution of hashtag use to the stance detection task, we have also used the existence of hashtags in tweets as an additional feature to unigrams. Hence, we take the classifiers using unigram-based features as the baseline systems. Evaluation results of the SVM classifiers using unigrams$+$hashtags as features are provided in Table \ref{tab:result2}.

\begin{table}[h!]
\centering
\caption{Evaluation Results of the SVM Classifiers Utilizing Unigrams$+$Hashtag Use as Features}
\label{tab:result2}
\begin{tabular}{|l|l|c|c|c|c|c|c|c|c|c|}
\hline
&&\multicolumn{9}{|c|}{Stance Data Set} \\
\cline{3-11}
\multirow{3}{*}{Target}&\multirow{3}{*}{Class}&\multicolumn{3}{|c|}{Version-1}&\multicolumn{3}{|c|}{Version-2}&\multicolumn{3}{|c|}{Version-3} \\
\cline{3-11}
&& P (\%) & R (\%) & F (\%) & P (\%) & R (\%) & F (\%) & P (\%) & R (\%) & F (\%) \\
\hline
\multirow{3}{*}{Target-1} & Favor & 75.0 & 90.9 & 82.2 & 73.7 & 89.0 & 80.6 & 74.3 & 93.3 & 82.7  \\
& Against & 88.4 & 69.7 & 78.0 & 86.1 & 68.2 & 76.1 & 91.0 & 67.5 & 77.5  \\
& \textbf{Average} & \textbf{81.7} & \textbf{80.3} & \textbf{80.1} & \textbf{79.9} & \textbf{78.6} & \textbf{78.4} & \textbf{82.6} & \textbf{80.4} & \textbf{80.1} \\
\hline
\multirow{3}{*}{Target-2} & Favor & 70.0 & 85.1 & 76.8  & 71.9 & 90.2 & 80.0 & 77.5 & 88.1 & 82.4  \\
& Against & 81.0 & 63.4 & 71.2 & 86.2 & 63.5 & 73.1 & 85.6 & 73.4 & 79.0  \\
& \textbf{Average} & \textbf{75.5} & \textbf{74.3} & \textbf{74.0} & \textbf{78.9} & \textbf{77.1} & \textbf{76.6} & \textbf{81.4} & \textbf{80.9} & \textbf{80.8} \\
\hline
\end{tabular}
\end{table}

When the results in Table \ref{tab:result2} are compared with the results in Table \ref{tab:result1}, for both targets, F-Measure rates have generally increased when hashtag use is also incorporated as an additional feature. That is, increases in the F-Measure rates are observed in five of the six target and data set version combinations. Only a decrease of 0.5\% is observed for \emph{Target-1} on the first version of the data set. The increases in the performance of both SVM classifiers, especially on the largest third version of the data set, constitute an encouraging evidence for the contribution of using features related to hashtags in a stance detection system. We leave other ways of exploiting hashtags for stance detection as future work.

We have also tested the contribution of other features on the top of the settings based on unigram$+$hashtag features. These additional features are:
\begin{itemize}
    \item   the existence of links to external Web sites
    \item   the existence of positive emoticons (such as \texttt{:)}, \texttt{:D}, \texttt{<3})
    \item   the existence of negative emoticons (such as \texttt{:(}, \texttt{:$\backslash$})
\end{itemize}

Our stance-detection experiments with SVM classifiers also utilizing these features do not lead to improved results. The results obtained using these features, either individually or together with the settings using unigram$+$hashtag features, are all slightly lower than the performance rates given in Table \ref{tab:result2}. Therefore, we can conclude that the use of features based on links and emoticons do not contribute to stance detection on our data sets, beyond the settings using unigram$+$hashtag features. Yet, we should note that especially the emoticon dictionaries that we have used are of limited size, therefore larger dictionaries can be employed and the related experiments can be repeated as part of future studies.

Overall, our performance evaluation results with SVMs given in this section are significant as reference results to be used for comparison purposes. They provide evidence for the utility of unigram-based and hashtag-related features in SVM classifiers for stance detection in Turkish tweets. The use of other features based on external links and emoticons have not improved the stance detection performance on our data sets, over the existing SVM settings with unigram$+$hashtag features.

\subsection{Experiments Using Named Entities as Features}\label{experimentsTr2}

In this section, we present the results of our stance detection experiments on using named entities as additional features. The idea of using named entities as stance detection features and the results of our related experiments on the first version (Version-1) of the data set have previously been presented in a preprint of us \cite{Kucuk2017joint}. Hence, this section includes and is built upon some of the content given in \cite{Kucuk2017joint}.

Named entity recognition (NER) has long been studied on formal text types such as news articles. Recently, due to the need for customized text processing approaches and tools for social media content such as tweets, considerable research effort has been devoted to NER on tweets \cite{ritter2011named,liu2011recognizing,Kucuk2014_2,derczynski2015analysis}. One observation commonly acknowledged in these studies is that the existing approaches performing favorably on well-formed texts perform poorly on tweets, due to the peculiarities of Twitter language. These peculiarities include using nicknames, contracted forms, instances of hypocorism and\slash or neologism, in addition to common writing\slash grammatical errors.

In the first subsection below, we describe our NER experiments on the stance data sets. In the following subsection, we present the results of our stance detection experiments on the data sets using named entities (both automatically-extracted ones and the ones in the annotated answer keys) as additional features. The results are accompanied with the discussions of the results for each experiment type and settings.

\subsubsection{Named Entity Recognition Experiments on the Stance Data Sets}

In order to carry out NER experiments on the three versions of our stance data sets, we have manually annotated them to use them as the answer keys. The named entity types considered are person, location, and organization names. Statistical information regarding these named entity annotations are provided in Table \ref{tab:ner_stats}, for each stance class and named entity type considered. The most common named entity type is organization names as expected, since each tweet contains the stance target (i.e., the football club name) explicitly and they are named entities of organization name type.

\begin{table}[h!]
\centering
\caption{Statistical Information on the Named Entities Annotated in the Three Versions of the Stance-Annotated Tweet Data Sets.}
\label{tab:ner_stats}
\begin{tabular}{|l|l|l|c|c|c|c|c|}
\hline
&&&&\multicolumn{4}{|c|}{\# of Named Entities} \\
\cline{5-8}
Stance Data Set & Target & Stance Class & \# of Tweets & Person & Location & Organization & \textbf{Total} \\
\hline
\multirow{5}{*}{Version-1} & \multirow{2}{*}{Target-1} & Favor & 175 & 12 & 17 & 207 & 236 \\
\cline{3-8}
                            && Against & 175 & 70 & 4 & 221 & 295 \\
\cline{2-8}
& \multirow{2}{*}{Target-2} & Favor & 175 & 8 & 24 & 247 & 279 \\
\cline{3-8}
                            && Against & 175 & 69 & 18 & 277 & 364 \\
\cline{2-8}
&\multicolumn{2}{|r|}{\textbf{Total}} & \textbf{700} & \textbf{159} & \textbf{63} & \textbf{952} & \textbf{1,174}\\
\hline
\multirow{5}{*}{Version-2} & \multirow{2}{*}{Target-1} & Favor & 173 & 11 & 14 & 205 & 230 \\
\cline{3-8}
                            && Against & 173 & 70 & 4 & 218 & 292 \\
\cline{2-8}
& \multirow{2}{*}{Target-2} & Favor & 173 & 6 & 24 & 245 & 275 \\
\cline{3-8}
                            && Against & 167 & 66 & 17 & 266 & 349 \\
\cline{2-8}
&\multicolumn{2}{|r|}{\textbf{Total}} & \textbf{686} & \textbf{153} & \textbf{59} & \textbf{934} & \textbf{1,146}\\
\hline
\multirow{5}{*}{Version-3} & \multirow{2}{*}{Target-1} & Favor & 269 & 29 & 25 & 336 & \textbf{390} \\
\cline{3-8}
                            && Against & 268 & 118 & 12 & 348 & 478 \\
\cline{2-8}
& \multirow{2}{*}{Target-2} & Favor & 269 & 20 & 34 & 411 & 465 \\
\cline{3-8}
                            && Against & 259 & 95 & 25 & 426 & 546 \\
\cline{2-8}
&\multicolumn{2}{|r|}{\textbf{Total}} & \textbf{1,065} & \textbf{262} & \textbf{96} & \textbf{1,521} & \textbf{1,879}\\
\hline
\end{tabular}
\end{table}

During our NER experiments on the three versions of the data sets, we have used the NER tool presented in \cite{Kucuk2014_2} for tweets, which, in turn, is the extended version of the rule-based recognizer for Turkish previously proposed in \cite{Kucuk2009} for news articles. Two lines of extensions over the original recognizer are: (1) relaxing the capitalization constraint of the tool to improve the NER performance on tweets where capitalizing names is often ignored, (2) diacritics-based expansion of the lexical resources of the tweets, as diacritics are often missed in names in tweets \cite{Kucuk2014_2}.

The evaluation results of this tool on the data sets are presented in Table \ref{tab:ner_results}, in terms of the common metrics of precision, recall, and F-Measure. During the calculation of these metrics, credit is given if both the boundaries and the type of the named entity extracted are true, i.e., no credit is given to partial extractions.

\begin{table}[h!]
\centering
\caption{Evaluation Results of the NER Tool on the Stance-Annotated Tweet Data Sets.}
\label{tab:ner_results}
\begin{tabular}{|l|l|l|c|c|c|}
\hline
Stance Data Set & Target & Stance Class & P (\%) & R (\%) & F (\%) \\
\hline
\multirow{4}{*}{Version-1} & \multirow{2}{*}{Target-1} & Favor & 73.79 & 64.41 & 68.78 \\
\cline{3-6}
                            && Against & 77.14 & 36.61 & 49.66 \\
\cline{2-6}
&\multirow{2}{*}{Target-2} & Favor & 78.80 & 51.97 & 62.63 \\
\cline{3-6}
                            && Against & 71.01 & 40.38 & 51.49 \\
\hline \hline
\multirow{4}{*}{Version-2} & \multirow{2}{*}{Target-1} & Favor & 73.37 & 63.48 & 68.07 \\
\cline{3-6}
                            && Against &  76.64 & 35.96 & 48.95 \\
\cline{2-6}
&\multirow{2}{*}{Target-2} & Favor &  79.12 & 52.36 & 63.02 \\
\cline{3-6}
                            && Against &  71.14 & 40.97 & 52.00 \\
\hline \hline
\multirow{4}{*}{Version-3} & \multirow{2}{*}{Target-1} & Favor &  72.79 & 57.22 & 64.07 \\
\cline{3-6}
                            && Against &  69.20 & 32.43 & 44.16 \\
\cline{2-6}
&\multirow{2}{*}{Target-2} & Favor &  77.42 & 46.45 & 58.06 \\
\cline{3-6}
                            && Against &  68.28 & 38.64 & 49.36 \\
\hline
\end{tabular}
\end{table}

The results presented in \ref{tab:ner_stats} are not very high considering the fact that the recognizer attains circa 85\% in F-Measure on news articles. Yet, they are favorable because the performance of NER tools is known to decrease considerably when ported to tweets \cite{ritter2011named} and the best F-Measure rate obtained by this extended NER tool on a randomly-compiled tweet data set is 48.13\% \cite{Kucuk2014_2}. The following two conclusions can be drawn based on these results:

\begin{itemize}
    \item   Overall, the results are better than those NER results reported on the randomly-compiled tweet sets. This may be attributed to the fact that all of the tweets in the three versions of our stance data sets contain the stance targets (the club names) explicitly, making our procedure "targeted" NER. That is, although some of them may be contracted forms or neologisms, the tweets contain the target names as organization names explicitly, as revealed with the large number of organization names in Table \ref{tab:ner_stats}, and a large proportion of them can be extracted by the NER tool \cite{Kucuk2017joint}.
    \item   The NER performance is far better on tweets marked with the \emph{Favor} stance class when compared with the performance on tweets marked as \emph{Against}. This observation can be due to the more common use of neologisms (for target names and in the negative sense) in tweets from the latter group than in tweets from the former group and these neologisms seem to be missed by the recognizer \cite{Kucuk2017joint}. The named entities in the tweets from the \emph{Favor} seem to be commonly well-formed and canonical names of these entities. The considerable difference in the recall rates for these two groups supports this argument.
\end{itemize}

The NER experiments given in this subsection are significant since: (i) NER research on tweets is still an important research problem in social media analysis and they are "targeted" NER experiments on a low-resource language, i.e., Turkish, (ii) the corresponding results and discussions may help reveal the interrelationships between stance annotation and named entity annotation. In the following subsection, the stance detection experiments using named entities as features are presented. We should note that, in Turkish, the uninflected named entities extracted and annotated in these data sets are no different than the corresponding ngrams, however, when they are inflected, the set of suffixes attached to the named entity are left out, both by the NER tool and during manual annotation to create the answer keys \cite{Kucuk2017joint}. Hence, inflected named entities in the stance data sets are different than the corresponding ngrams, making these entities plausible features for stance detection, as discussed in the following section.

\subsubsection{Stance Detection Experiments Using Named Entities as Features}

In the previous sections of this paper, our experiment results demonstrate that features based on unigrams and hashtag use can be used to improve stance detection performance. In this section, we present the evaluation results of using named entities as additional features for stance detection without considering the types of these named entities.

Firstly, in Table \ref{tab:ner_stance_result_system}, we present the performance results of stance detection experiments using the automatically-extracted named entities as features, together with unigrams and hashtag use. These evaluation results are all better than those corresponding results in Table \ref{tab:result2} where unigrams and hashtag use are employed as the sole set of features. Hence, we can conclude that the performance rates are improved with the incorporation of named entities as new features, even when these named entities are extracted through an automated procedure.

\begin{table}[h!]
\centering
\caption{Evaluation Results of the SVM Classifiers Utilizing Unigrams$+$Hashtag Use$+$Named Entities as Features, with Named Entities Extracted by the NER Tool.}
\label{tab:ner_stance_result_system}
\begin{tabular}{|l|l|c|c|c|c|c|c|c|c|c|}
\hline
&&\multicolumn{9}{|c|}{Stance Data Set} \\
\cline{3-11}
\multirow{3}{*}{Target}&\multirow{3}{*}{Class}&\multicolumn{3}{|c|}{Version-1}&\multicolumn{3}{|c|}{Version-2}&\multicolumn{3}{|c|}{Version-3} \\
\cline{3-11}
&& P (\%) & R (\%) & F (\%) & P (\%) & R (\%) & F (\%) & P (\%) & R (\%) & F (\%) \\
\hline
\multirow{3}{*}{Target-1} & Favor & 75.6 & 90.3 & 82.3 & 74.5 & 87.9 & 80.6 & 77.5  & 93.3  & 84.7  \\
& Against & 87.9 & 70.9 & 78.5 & 85.2 & 69.9 & 76.8 &  91.5 & 72.8  & 81.1    \\
& \textbf{Average} & \textbf{81.8} & \textbf{80.6} & \textbf{80.4} & \textbf{79.9} & \textbf{78.9} & \textbf{78.7} & \textbf{84.5} & \textbf{83.1} & \textbf{82.9} \\
\hline
\multirow{3}{*}{Target-2} & Favor & 71.8 & 84.6 & 77.7 & 73.9 & 93.1 & 82.4 & 78.6 & 90.0 & 83.9  \\
& Against & 81.3 & 66.9 & 73.4 & 90.2 & 65.9 & 76.1 & 87.7 & 74.5 & 80.6  \\
& \textbf{Average} & \textbf{76.5} & \textbf{75.7} & \textbf{75.5} & \textbf{81.9} & \textbf{79.7} & \textbf{79.3} & \textbf{83.1} & \textbf{82.4} & \textbf{82.3} \\
\hline
\end{tabular}
\end{table}

Similarly, Table \ref{tab:ner_stance_result_answer} presents the evaluation results utilizing the named entities from the manually-annotated answer key as additional features, together with unigrams and hashtag use. These results are more favorable than those corresponding results in Table \ref{tab:ner_stance_result_system} which indicates stance detection performance on Turkish tweets can be further improved with the use of named entities extracted with high precision and recall.

The findings provided in this section reveal that the use of named entities is an improving feature for the stance detection task on Turkish tweets. They also imply that the interrelationship between the well-studied topic of NER and the newly-emerged topic of stance detection is a fruitful research topic which requires further research attention.

\begin{table}[h!]
\centering
\caption{Evaluation Results of the SVM Classifiers Utilizing Unigrams$+$Hashtag Use$+$Named Entities as Features, with Named Entities from the Manually-Annotated Answer Key.}
\label{tab:ner_stance_result_answer}
\begin{tabular}{|l|l|c|c|c|c|c|c|c|c|c|}
\hline
&&\multicolumn{9}{|c|}{Stance Data Set} \\
\cline{3-11}
\multirow{3}{*}{Target}&\multirow{3}{*}{Class}&\multicolumn{3}{|c|}{Version-1}&\multicolumn{3}{|c|}{Version-2}&\multicolumn{3}{|c|}{Version-3} \\
\cline{3-11}
&& P (\%) & R (\%) & F (\%) & P (\%) & R (\%) & F (\%) & P (\%) & R (\%) & F (\%) \\
\hline
\multirow{3}{*}{Target-1} & Favor & 76.3 & 92.0 & 83.4 & 0.0 & 0.0 & 0.0 & 78.8 & 94.1 & 85.8  \\
& Against & 89.9 & 71.4 & 79.6 & 0.0 & 0.0 & 0.0 & 92.6 & 74.6 & 82.6  \\
& \textbf{Average} & \textbf{83.1} & \textbf{81.7} & \textbf{81.5} & \textbf{0.0} & \textbf{0.0} & \textbf{0.0} & \textbf{85.7} & \textbf{84.4} & \textbf{84.2} \\
\hline
\multirow{3}{*}{Target-2} & Favor & 74.3 & 90.9 & 81.7 & 0.0 & 0.0 & 0.0 & 78.4 & 91.8 & 84.6  \\
& Against & 88.2 & 68.6 & 77.2 & 0.0 & 0.0 & 0.0 & 89.7 & 73.7 & 80.9  \\
& \textbf{Average} & \textbf{81.3} & \textbf{79.7} & \textbf{79.5} & \textbf{0.0} & \textbf{0.0} & \textbf{0.0} & \textbf{83.9} & \textbf{83.0} & \textbf{82.8} \\
\hline
\end{tabular}
\end{table}

\newpage

\section{Future Work}\label{future}
Below listed are future research directions based on the current study:
\begin{itemize}
  \item Within the course of further extensions and revisions on the stance detection data set versions, other stance classes like \emph{Neither} can also be considered to make them compatible with the data sets in English proposed in the literature.
  \item Our results in the current paper suggest that the existence of hashtags is boosting feature for stance detection in Turkish tweets. Other ways of utilizing hashtags for stance detection can be considered, which may require deeper processing of these hashtags.
  \item Our evaluation results also indicate that named entities can be improving factors for stance detection. Similar experiments on tweets in other languages can be conducted so that based on the corresponding results, more general conclusions can be drawn.
  \item Other features employed for stance detection in the literature can also be used by the classifiers and these classifiers can be tested on larger data sets. For instance, larger emoticon dictionaries can be utilized during the employment of emoticon-based features. Other classification approaches presented in recent studies such as \cite{mohammad2016semeval} could also be tested against our baseline classifiers for comparison purposes.
  \item Methods and resources (such as polarity lexicons) used for sentiment analysis can also be employed to improve the stance detection performance on tweets, as part of future work.
  \item Prior to stance detection on tweets, a tweet normalization procedure \cite{han2011lexical} can be employed, in order to determine whether tweet normalization is a facilitating or an impeding factor for stance detection in tweets.
  \item Lastly, the SVM classifiers utilized in this study and their prospective versions utilizing different feature sets can be tested on stance data sets in other languages for comparison purposes.
\end{itemize}

\section{Conclusion}\label{conc}
Stance detection is a considerably new research area in NLP, which is a particular subproblem of the well-studied topic of sentiment analysis or opinion mining. The aim of the stance detection approaches is to identify the stance, or position, towards a target which may be explicitly specified in the text or not. In this study, we propose different versions of a stance-annotated sports-related tweet data set in Turkish and present the performance evaluation results of SVM classifiers for stance detection on these versions. The data sets comprise tweets regarding two well-known football clubs which constitute the stance targets and the corresponding stance annotations are made publicly available. The sets are quite significant as there is a scarcity of annotated resources for NLP tasks in non-English languages and to the best of our knowledge, this is the first stance-annotated data set for Turkish and also the first sports-related one in the general sense.

On the other hand, we have carried out experiments with SVM classifiers for each target with various features including those based on unigrams, bigrams, hashtags, external links, positive and negative emoticons, and named entities. The 10-fold cross validation results of the SVM classifiers using unigrams$+$hashtag use$+$named entities have been promising and better than the other combinations tested. Therefore, our findings suggest that features based on unigrams, hashtags, and named entities can be used to improve stance detection performance on sports-related tweets.

\bibliographystyle{ACM-Reference-Format}

\begin{thebibliography}{00}


\ifx \showCODEN    \undefined \def \showCODEN     #1{\unskip}     \fi
\ifx \showDOI      \undefined \def \showDOI       #1{#1}\fi
\ifx \showISBNx    \undefined \def \showISBNx     #1{\unskip}     \fi
\ifx \showISBNxiii \undefined \def \showISBNxiii  #1{\unskip}     \fi
\ifx \showISSN     \undefined \def \showISSN      #1{\unskip}     \fi
\ifx \showLCCN     \undefined \def \showLCCN      #1{\unskip}     \fi
\ifx \shownote     \undefined \def \shownote      #1{#1}          \fi
\ifx \showarticletitle \undefined \def \showarticletitle #1{#1}   \fi
\ifx \showURL      \undefined \def \showURL       {\relax}        \fi
\providecommand\bibfield[2]{#2}
\providecommand\bibinfo[2]{#2}
\providecommand\natexlab[1]{#1}
\providecommand\showeprint[2][]{arXiv:#2}

\bibitem[\protect\citeauthoryear{Augenstein, Rockt{\"a}schel, Vlachos, and
  Bontcheva}{Augenstein et~al\mbox{.}}{2016}]%
        {augenstein2016stance}
\bibfield{author}{\bibinfo{person}{Isabelle Augenstein}, \bibinfo{person}{Tim
  Rockt{\"a}schel}, \bibinfo{person}{Andreas Vlachos}, {and}
  \bibinfo{person}{Kalina Bontcheva}.} \bibinfo{year}{2016}\natexlab{}.
\newblock \showarticletitle{Stance detection with bidirectional conditional
  encoding}. In \bibinfo{booktitle}{{\em Proceedings of the Conference on
  Empirical Methods in Natural Language Processing}}.
  \bibinfo{pages}{876--885}.
\newblock


\bibitem[\protect\citeauthoryear{Can, Kocberber, Balcik, Kaynak, Ocalan, and
  Vursavas}{Can et~al\mbox{.}}{2008}]%
        {can2008information}
\bibfield{author}{\bibinfo{person}{Fazli Can}, \bibinfo{person}{Seyit
  Kocberber}, \bibinfo{person}{Erman Balcik}, \bibinfo{person}{Cihan Kaynak},
  \bibinfo{person}{H~Cagdas Ocalan}, {and} \bibinfo{person}{Onur~M Vursavas}.}
  \bibinfo{year}{2008}\natexlab{}.
\newblock \showarticletitle{Information retrieval on Turkish texts}.
\newblock \bibinfo{journal}{{\em Journal of the American Society for
  Information Science and Technology\/}} \bibinfo{volume}{59},
  \bibinfo{number}{3}, \bibinfo{pages}{407--421}.
\newblock


\bibitem[\protect\citeauthoryear{Chen and Chen}{Chen and Chen}{2016}]%
        {chen2016scifnet}
\bibfield{author}{\bibinfo{person}{Zhong-Yong Chen} {and}
  \bibinfo{person}{Chien~Chin Chen}.} \bibinfo{year}{2016}\natexlab{}.
\newblock \showarticletitle{SCIFNET: Stance community identification of topic
  persons using friendship network analysis}.
\newblock \bibinfo{journal}{{\em Knowledge-Based Systems\/}}
  \bibinfo{volume}{110}, \bibinfo{pages}{30--48}.
\newblock


\bibitem[\protect\citeauthoryear{Cohen}{Cohen}{1960}]%
        {cohen1960coefficient}
\bibfield{author}{\bibinfo{person}{Jacob Cohen}.}
  \bibinfo{year}{1960}\natexlab{}.
\newblock \showarticletitle{A coefficient of agreement for nominal scales}.
\newblock \bibinfo{journal}{{\em Educational and Psychological Measurement\/}}
  \bibinfo{volume}{20}, \bibinfo{number}{1}, \bibinfo{pages}{37--46}.
\newblock


\bibitem[\protect\citeauthoryear{Derczynski, Maynard, Rizzo, van Erp, Gorrell,
  Troncy, Petrak, and Bontcheva}{Derczynski et~al\mbox{.}}{2015}]%
        {derczynski2015analysis}
\bibfield{author}{\bibinfo{person}{Leon Derczynski}, \bibinfo{person}{Diana
  Maynard}, \bibinfo{person}{Giuseppe Rizzo}, \bibinfo{person}{Marieke van
  Erp}, \bibinfo{person}{Genevieve Gorrell}, \bibinfo{person}{Rapha{\"e}l
  Troncy}, \bibinfo{person}{Johann Petrak}, {and} \bibinfo{person}{Kalina
  Bontcheva}.} \bibinfo{year}{2015}\natexlab{}.
\newblock \showarticletitle{Analysis of named entity recognition and linking
  for tweets}.
\newblock \bibinfo{journal}{{\em Information Processing \& Management\/}}
  \bibinfo{volume}{51}, \bibinfo{number}{2}, \bibinfo{pages}{32--49}.
\newblock


\bibitem[\protect\citeauthoryear{Dori-Hacohen}{Dori-Hacohen}{2015}]%
        {dori2015controversy}
\bibfield{author}{\bibinfo{person}{Shiri Dori-Hacohen}.}
  \bibinfo{year}{2015}\natexlab{}.
\newblock \showarticletitle{Controversy detection and stance analysis}. In
  \bibinfo{booktitle}{{\em Proceedings of the International ACM Conference on
  Research and Development in Information Retrieval}}. \bibinfo{pages}{1057}.
\newblock


\bibitem[\protect\citeauthoryear{Ebrahimi, Dou, and Lowd}{Ebrahimi
  et~al\mbox{.}}{2016}]%
        {ebrahimijoint2016}
\bibfield{author}{\bibinfo{person}{Javid Ebrahimi}, \bibinfo{person}{Dejing
  Dou}, {and} \bibinfo{person}{Daniel Lowd}.} \bibinfo{year}{2016}\natexlab{}.
\newblock \showarticletitle{A joint sentiment-target-stance model for stance
  classification in tweets}. In \bibinfo{booktitle}{{\em Proceedings of the
  International Conference on Computational Linguistics}}.
  \bibinfo{pages}{2656--2665}.
\newblock


\bibitem[\protect\citeauthoryear{Faulkner}{Faulkner}{2014}]%
        {faulkner2014automated}
\bibfield{author}{\bibinfo{person}{Adam Faulkner}.}
  \bibinfo{year}{2014}\natexlab{}.
\newblock \showarticletitle{Automated classification of stance in student
  essays: An approach using stance target information and the Wikipedia
  link-based measure}. In \bibinfo{booktitle}{{\em Proceedings of the
  International Florida Artificial Intelligence Research Society Conference}}.
  \bibinfo{pages}{174--179}.
\newblock


\bibitem[\protect\citeauthoryear{Gadek, Betsholtz, Pauchet, Brunessaux,
  Malandain, and Vercouter}{Gadek et~al\mbox{.}}{2017}]%
        {gadek2017}
\bibfield{author}{\bibinfo{person}{Guillaume Gadek}, \bibinfo{person}{Josefin
  Betsholtz}, \bibinfo{person}{Alexandre Pauchet},
  \bibinfo{person}{St{\'{e}}phan Brunessaux}, \bibinfo{person}{Nicolas
  Malandain}, {and} \bibinfo{person}{Laurent Vercouter}.}
  \bibinfo{year}{2017}\natexlab{}.
\newblock \showarticletitle{Extracting contextonyms from Twitter for stance
  detection}. In \bibinfo{booktitle}{{\em Proceedings of the International
  Conference on Agents and Artificial Intelligence}}.
  \bibinfo{pages}{132--141}.
\newblock


\bibitem[\protect\citeauthoryear{Hall, Frank, Holmes, Pfahringer, Reutemann,
  and Witten}{Hall et~al\mbox{.}}{2009}]%
        {weka2009}
\bibfield{author}{\bibinfo{person}{Mark Hall}, \bibinfo{person}{Eibe Frank},
  \bibinfo{person}{Geoffrey Holmes}, \bibinfo{person}{Bernhard Pfahringer},
  \bibinfo{person}{Peter Reutemann}, {and} \bibinfo{person}{Ian~H Witten}.}
  \bibinfo{year}{2009}\natexlab{}.
\newblock \showarticletitle{The WEKA data mining software: An update}.
\newblock \bibinfo{journal}{{\em ACM SIGKDD Explorations Newsletter\/}}
  \bibinfo{volume}{11}, \bibinfo{number}{1}, \bibinfo{pages}{10--18}.
\newblock


\bibitem[\protect\citeauthoryear{Han and Baldwin}{Han and Baldwin}{2011}]%
        {han2011lexical}
\bibfield{author}{\bibinfo{person}{Bo Han} {and} \bibinfo{person}{Timothy
  Baldwin}.} \bibinfo{year}{2011}\natexlab{}.
\newblock \showarticletitle{Lexical normalisation of short text messages: Makn
  sens a \#twitter}. In \bibinfo{booktitle}{{\em Proceedings of the 49th Annual
  Meeting of the Association for Computational Linguistics: Human Language
  Technologies-Volume 1}}. \bibinfo{pages}{368--378}.
\newblock


\bibitem[\protect\citeauthoryear{Hasan and Ng}{Hasan and Ng}{2013}]%
        {hasan2013stance}
\bibfield{author}{\bibinfo{person}{Kazi~Saidul Hasan} {and}
  \bibinfo{person}{Vincent Ng}.} \bibinfo{year}{2013}\natexlab{}.
\newblock \showarticletitle{Stance classification of ideological debates: Data,
  models, features, and constraints}. In \bibinfo{booktitle}{{\em Proceedings
  of the International Joint Conference on Natural Language Processing}}.
  \bibinfo{pages}{1348--1356}.
\newblock


\bibitem[\protect\citeauthoryear{K\"u\c{c}\"uk}{K\"u\c{c}\"uk}{2011}]%
        {kucuk2011exploiting}
\bibfield{author}{\bibinfo{person}{Dilek K\"u\c{c}\"uk}.}
  \bibinfo{year}{2011}\natexlab{}.
\newblock {\em \bibinfo{title}{Exploiting information extraction techniques for
  automatic semantic annotation and retrieval of news videos in Turkish}}.
\newblock \bibinfo{thesistype}{Ph.D. Dissertation}. \bibinfo{school}{Middle
  East Technical University}.
\newblock


\bibitem[\protect\citeauthoryear{K\"u\c{c}\"uk}{K\"u\c{c}\"uk}{2017a}]%
        {Kucuk2017joint}
\bibfield{author}{\bibinfo{person}{Dilek K\"u\c{c}\"uk}.}
  \bibinfo{year}{2017}\natexlab{a}.
\newblock \showarticletitle{Joint named entity recognition and stance detection
  in tweets}.
\newblock \bibinfo{journal}{{\em arXiv preprint arXiv:1707.09611\/}}.
\newblock


\bibitem[\protect\citeauthoryear{K\"u\c{c}\"uk}{K\"u\c{c}\"uk}{2017b}]%
        {kucuk2017stance}
\bibfield{author}{\bibinfo{person}{Dilek K\"u\c{c}\"uk}.}
  \bibinfo{year}{2017}\natexlab{b}.
\newblock \showarticletitle{Stance detection in Turkish tweets}. In
  \bibinfo{booktitle}{{\em Proceedings of the International Workshop on Social
  Media World Sensors}}.
\newblock


\bibitem[\protect\citeauthoryear{K\"u\c{c}\"uk and Steinberger}{K\"u\c{c}\"uk
  and Steinberger}{2014}]%
        {Kucuk2014_2}
\bibfield{author}{\bibinfo{person}{Dilek K\"u\c{c}\"uk} {and}
  \bibinfo{person}{Ralf Steinberger}.} \bibinfo{year}{2014}\natexlab{}.
\newblock \showarticletitle{Experiments to improve named entity recognition on
  {Turkish} tweets}. In \bibinfo{booktitle}{{\em Proceedings of the EACL
  Workshop on Language Analysis for Social Media}}. \bibinfo{pages}{71--78}.
\newblock


\bibitem[\protect\citeauthoryear{K\"u\c{c}\"uk and Yaz{\i}c{\i}}{K\"u\c{c}\"uk
  and Yaz{\i}c{\i}}{2009}]%
        {Kucuk2009}
\bibfield{author}{\bibinfo{person}{Dilek K\"u\c{c}\"uk} {and}
  \bibinfo{person}{Adnan Yaz{\i}c{\i}}.} \bibinfo{year}{2009}\natexlab{}.
\newblock \showarticletitle{Named entity recognition experiments on {Turkish}
  texts}.
\newblock In \bibinfo{booktitle}{{\em Proceedings of the International
  Conference on Flexible Query Answering Systems}}. \bibinfo{series}{LNCS},
  Vol.~\bibinfo{volume}{5822}. \bibinfo{pages}{524--535}.
\newblock


\bibitem[\protect\citeauthoryear{K\"u\c{c}\"uk, Yapar, K\"u\c{c}\"uk, and
  K\"u\c{c}\"uk}{K\"u\c{c}\"uk et~al\mbox{.}}{2017}]%
        {Kucuk2017}
\bibfield{author}{\bibinfo{person}{Emine~Ela K\"u\c{c}\"uk},
  \bibinfo{person}{K\"ur\c{s}ad Yapar}, \bibinfo{person}{Dilek K\"u\c{c}\"uk},
  {and} \bibinfo{person}{Do\u{g}an K\"u\c{c}\"uk}.}
  \bibinfo{year}{2017}\natexlab{}.
\newblock \showarticletitle{Ontology-based automatic identification of public
  health-related {Turkish} tweets}.
\newblock \bibinfo{journal}{{\em Computers in Biology and Medicine\/}}
  \bibinfo{volume}{83}, \bibinfo{pages}{1--9}.
\newblock


\bibitem[\protect\citeauthoryear{Liu, Zhang, Wei, and Zhou}{Liu
  et~al\mbox{.}}{2011}]%
        {liu2011recognizing}
\bibfield{author}{\bibinfo{person}{Xiaohua Liu}, \bibinfo{person}{Shaodian
  Zhang}, \bibinfo{person}{Furu Wei}, {and} \bibinfo{person}{Ming Zhou}.}
  \bibinfo{year}{2011}\natexlab{}.
\newblock \showarticletitle{Recognizing named entities in tweets}. In
  \bibinfo{booktitle}{{\em Proceedings of the Annual Meeting of the Association
  for Computational Linguistics: Human Language Technologies-Volume 1}}.
  \bibinfo{pages}{359--367}.
\newblock


\bibitem[\protect\citeauthoryear{Misra and Walker}{Misra and Walker}{2013}]%
        {misra2013topic}
\bibfield{author}{\bibinfo{person}{Amita Misra} {and}
  \bibinfo{person}{Marilyn~A Walker}.} \bibinfo{year}{2013}\natexlab{}.
\newblock \showarticletitle{Topic independent identification of agreement and
  disagreement in social media dialogue}. In \bibinfo{booktitle}{{\em
  Proceedings of the Conference of the Special Interest Group on Discourse and
  Dialogue}}. \bibinfo{pages}{41--50}.
\newblock


\bibitem[\protect\citeauthoryear{Mohammad, Kiritchenko, Sobhani, Zhu, and
  Cherry}{Mohammad et~al\mbox{.}}{2016a}]%
        {mohammad2016dataset}
\bibfield{author}{\bibinfo{person}{Saif~M Mohammad}, \bibinfo{person}{Svetlana
  Kiritchenko}, \bibinfo{person}{Parinaz Sobhani}, \bibinfo{person}{Xiaodan
  Zhu}, {and} \bibinfo{person}{Colin Cherry}.}
  \bibinfo{year}{2016}\natexlab{a}.
\newblock \showarticletitle{A dataset for detecting stance in tweets}. In
  \bibinfo{booktitle}{{\em Proceedings of the Language Resources and Evaluation
  Conference}}. \bibinfo{pages}{3945--3952}.
\newblock


\bibitem[\protect\citeauthoryear{Mohammad, Kiritchenko, Sobhani, Zhu, and
  Cherry}{Mohammad et~al\mbox{.}}{2016b}]%
        {mohammad2016semeval}
\bibfield{author}{\bibinfo{person}{Saif~M Mohammad}, \bibinfo{person}{Svetlana
  Kiritchenko}, \bibinfo{person}{Parinaz Sobhani}, \bibinfo{person}{Xiaodan
  Zhu}, {and} \bibinfo{person}{Colin Cherry}.}
  \bibinfo{year}{2016}\natexlab{b}.
\newblock \showarticletitle{SemEval-2016 task 6: Detecting stance in tweets}.
  In \bibinfo{booktitle}{{\em Proceedings of the International Workshop on
  Semantic Evaluation}}. \bibinfo{pages}{31--41}.
\newblock


\bibitem[\protect\citeauthoryear{Mohammad, Sobhani, and Kiritchenko}{Mohammad
  et~al\mbox{.}}{2017}]%
        {mohammad2017stance}
\bibfield{author}{\bibinfo{person}{Saif~M Mohammad}, \bibinfo{person}{Parinaz
  Sobhani}, {and} \bibinfo{person}{Svetlana Kiritchenko}.}
  \bibinfo{year}{2017}\natexlab{}.
\newblock \showarticletitle{Stance and sentiment in tweets}.
\newblock \bibinfo{journal}{{\em ACM Transactions on Internet Technology\/}}
  \bibinfo{volume}{17}, \bibinfo{number}{3}, \bibinfo{pages}{Article 26}.
\newblock


\bibitem[\protect\citeauthoryear{Nguyen and Litman}{Nguyen and Litman}{2015}]%
        {nguyen2015extracting}
\bibfield{author}{\bibinfo{person}{Huy~V Nguyen} {and} \bibinfo{person}{Diane~J
  Litman}.} \bibinfo{year}{2015}\natexlab{}.
\newblock \showarticletitle{Extracting argument and domain words for
  identifying argument components in texts}. In \bibinfo{booktitle}{{\em
  Proceedings of the Workshop on Argumentation Mining}}.
  \bibinfo{pages}{22--28}.
\newblock


\bibitem[\protect\citeauthoryear{Pang and Lee}{Pang and Lee}{2008}]%
        {pang2008opinion}
\bibfield{author}{\bibinfo{person}{Bo Pang} {and} \bibinfo{person}{Lillian
  Lee}.} \bibinfo{year}{2008}\natexlab{}.
\newblock \showarticletitle{Opinion mining and sentiment analysis}.
\newblock \bibinfo{journal}{{\em Foundations and Trends in Information
  Retrieval\/}} \bibinfo{volume}{2}, \bibinfo{number}{1-2},
  \bibinfo{pages}{1--135}.
\newblock


\bibitem[\protect\citeauthoryear{Platt}{Platt}{1999}]%
        {platt1999}
\bibfield{author}{\bibinfo{person}{John~C. Platt}.}
  \bibinfo{year}{1999}\natexlab{}.
\newblock \showarticletitle{Fast training of support vector machines using
  sequential minimal optimization}.
\newblock \bibinfo{journal}{{\em Advances in Kernel Methods\/}},
  \bibinfo{pages}{185--208}.
\newblock


\bibitem[\protect\citeauthoryear{Poursepanj, Weissbock, and Inkpen}{Poursepanj
  et~al\mbox{.}}{2013}]%
        {poursepanj2013uottawa}
\bibfield{author}{\bibinfo{person}{Hamid Poursepanj}, \bibinfo{person}{Josh
  Weissbock}, {and} \bibinfo{person}{Diana Inkpen}.}
  \bibinfo{year}{2013}\natexlab{}.
\newblock \showarticletitle{uOttawa: System description for SemEval-2013 task 2
  sentiment analysis in Twitter}. In \bibinfo{booktitle}{{\em Proceedings of
  the Joint Conference on Lexical and Computational Semantics (*SEM), Volume 2:
  International Workshop on Semantic Evaluation}}. \bibinfo{pages}{380--383}.
\newblock


\bibitem[\protect\citeauthoryear{Ritter, Clark, Etzioni, et~al\mbox{.}}{Ritter
  et~al\mbox{.}}{2011}]%
        {ritter2011named}
\bibfield{author}{\bibinfo{person}{Alan Ritter}, \bibinfo{person}{Sam Clark},
  \bibinfo{person}{Oren Etzioni}, {et~al\mbox{.}}}
  \bibinfo{year}{2011}\natexlab{}.
\newblock \showarticletitle{Named entity recognition in tweets: an experimental
  study}. In \bibinfo{booktitle}{{\em Proceedings of the Conference on
  Empirical Methods in Natural Language Processing}}.
  \bibinfo{pages}{1524--1534}.
\newblock


\bibitem[\protect\citeauthoryear{Sobhani, Inkpen, and Matwin}{Sobhani
  et~al\mbox{.}}{2015}]%
        {sobhani2015argumentation}
\bibfield{author}{\bibinfo{person}{Parinaz Sobhani}, \bibinfo{person}{Diana
  Inkpen}, {and} \bibinfo{person}{Stan Matwin}.}
  \bibinfo{year}{2015}\natexlab{}.
\newblock \showarticletitle{From argumentation mining to stance
  classification}. In \bibinfo{booktitle}{{\em Proceedings of the Workshop on
  Argumentation Mining}}. \bibinfo{pages}{67--77}.
\newblock


\bibitem[\protect\citeauthoryear{Sobhani, Inkpen, and Zhu}{Sobhani
  et~al\mbox{.}}{2017}]%
        {sobhani2017dataset}
\bibfield{author}{\bibinfo{person}{Parinaz Sobhani}, \bibinfo{person}{Diana
  Inkpen}, {and} \bibinfo{person}{Xiaodan Zhu}.}
  \bibinfo{year}{2017}\natexlab{}.
\newblock \showarticletitle{A dataset for multi-target stance detection}. In
  \bibinfo{booktitle}{{\em Proceedings of the Conference of the European
  Chapter of the Association for Computational Linguistics}}.
  \bibinfo{pages}{551--557}.
\newblock


\bibitem[\protect\citeauthoryear{Somasundaran and Wiebe}{Somasundaran and
  Wiebe}{2010}]%
        {somasundaran2010recognizing}
\bibfield{author}{\bibinfo{person}{Swapna Somasundaran} {and}
  \bibinfo{person}{Janyce Wiebe}.} \bibinfo{year}{2010}\natexlab{}.
\newblock \showarticletitle{Recognizing stances in ideological on-line
  debates}. In \bibinfo{booktitle}{{\em Proceedings of the Workshop on
  Computational Approaches to Analysis and Generation of Emotion in Text}}.
  \bibinfo{pages}{116--124}.
\newblock


\bibitem[\protect\citeauthoryear{Sridhar, Foulds, Huang, Getoor, and
  Walker}{Sridhar et~al\mbox{.}}{2015}]%
        {sridhar2015joint}
\bibfield{author}{\bibinfo{person}{Dhanya Sridhar}, \bibinfo{person}{James
  Foulds}, \bibinfo{person}{Bert Huang}, \bibinfo{person}{Lise Getoor}, {and}
  \bibinfo{person}{Marilyn Walker}.} \bibinfo{year}{2015}\natexlab{}.
\newblock \showarticletitle{Joint models of disagreement and stance in online
  debate}. In \bibinfo{booktitle}{{\em Proceedings of the Annual Meeting of the
  Association for Computational Linguistics and the International Joint
  Conference on Natural Language Processing}}. \bibinfo{pages}{116--125}.
\newblock


\bibitem[\protect\citeauthoryear{Tutek, Sekulic, Gombar, Paljak, Culinovic,
  Boltuzic, Karan, Alagi{\'c}, and {\v{S}}najder}{Tutek et~al\mbox{.}}{2016}]%
        {tutek2016takelab}
\bibfield{author}{\bibinfo{person}{Martin Tutek}, \bibinfo{person}{Ivan
  Sekulic}, \bibinfo{person}{Paula Gombar}, \bibinfo{person}{Ivan Paljak},
  \bibinfo{person}{Filip Culinovic}, \bibinfo{person}{Filip Boltuzic},
  \bibinfo{person}{Mladen Karan}, \bibinfo{person}{Domagoj Alagi{\'c}}, {and}
  \bibinfo{person}{Jan {\v{S}}najder}.} \bibinfo{year}{2016}\natexlab{}.
\newblock \showarticletitle{Takelab at SemEval-2016 task 6: Stance
  classification in tweets using a genetic algorithm based ensemble}. In
  \bibinfo{booktitle}{{\em Proceedings of the International Workshop on
  Semantic Evaluation}}. \bibinfo{pages}{464--468}.
\newblock


\bibitem[\protect\citeauthoryear{Walker, Anand, Abbott, and Grant}{Walker
  et~al\mbox{.}}{2012a}]%
        {walker2012stance}
\bibfield{author}{\bibinfo{person}{Marilyn~A Walker}, \bibinfo{person}{Pranav
  Anand}, \bibinfo{person}{Robert Abbott}, {and} \bibinfo{person}{Ricky
  Grant}.} \bibinfo{year}{2012}\natexlab{a}.
\newblock \showarticletitle{Stance classification using dialogic properties of
  persuasion}. In \bibinfo{booktitle}{{\em Proceedings of the Conference of the
  North American Chapter of the Association for Computational Linguistics:
  Human Language Technologies}}. \bibinfo{pages}{592--596}.
\newblock


\bibitem[\protect\citeauthoryear{Walker, Anand, Abbott, Tree, Martell, and
  King}{Walker et~al\mbox{.}}{2012b}]%
        {walker2012your}
\bibfield{author}{\bibinfo{person}{Marilyn~A Walker}, \bibinfo{person}{Pranav
  Anand}, \bibinfo{person}{Rob Abbott}, \bibinfo{person}{Jean E~Fox Tree},
  \bibinfo{person}{Craig Martell}, {and} \bibinfo{person}{Joseph King}.}
  \bibinfo{year}{2012}\natexlab{b}.
\newblock \showarticletitle{That is your evidence?: Classifying stance in
  online political debate}.
\newblock \bibinfo{journal}{{\em Decision Support Systems\/}}
  \bibinfo{volume}{53}, \bibinfo{number}{4}, \bibinfo{pages}{719--729}.
\newblock


\bibitem[\protect\citeauthoryear{Wei, Zhang, Liu, Chen, and Wang}{Wei
  et~al\mbox{.}}{2016}]%
        {wei2016pkudblab}
\bibfield{author}{\bibinfo{person}{Wan Wei}, \bibinfo{person}{Xiao Zhang},
  \bibinfo{person}{Xuqin Liu}, \bibinfo{person}{Wei Chen}, {and}
  \bibinfo{person}{Tengjiao Wang}.} \bibinfo{year}{2016}\natexlab{}.
\newblock \showarticletitle{pkudblab at SemEval-2016 Task 6: A specific
  convolutional neural network system for effective stance detection}. In
  \bibinfo{booktitle}{{\em Proceedings of the International Workshop on
  Semantic Evaluation}}. \bibinfo{pages}{384--388}.
\newblock


\bibitem[\protect\citeauthoryear{Zarrella and Marsh}{Zarrella and
  Marsh}{2016}]%
        {zarrella2016mitre}
\bibfield{author}{\bibinfo{person}{Guido Zarrella} {and} \bibinfo{person}{Amy
  Marsh}.} \bibinfo{year}{2016}\natexlab{}.
\newblock \showarticletitle{MITRE at SemEval-2016 task 6: Transfer learning for
  stance detection}. In \bibinfo{booktitle}{{\em Proceedings of the
  International Workshop on Semantic Evaluation}}. \bibinfo{pages}{458--463}.
\newblock


\end{thebibliography}

\end{document}